\documentclass{article}

\PassOptionsToPackage{numbers, compress}{natbib}
\usepackage{nips_2017}

\usepackage[utf8]{inputenc} % allow utf-8 input
\usepackage[T1]{fontenc}    % use 8-bit T1 fonts
\usepackage{hyperref}       % hyperlinks
\usepackage{url}            % simple URL typesetting
\usepackage{booktabs}       % professional-quality tables
\usepackage{amsfonts}       % blackboard math symbols
\usepackage{nicefrac}       % compact symbols for 1/2, etc.
\usepackage{microtype}      % microtypography
\usepackage{graphicx}
\usepackage{amsthm}
\usepackage{algorithm}
\usepackage{algorithmic}

\title{Deep Reinforcement Learning with Surrogate Agent-Environment Interface}

\author{
  Song Wang\\
  \texttt{songwang.academic@gmail.com} \\
  \And
  Yu Jing \\
  Wayne State University \\
  \texttt{yu.jing@wayne.edu} \\
}

\begin{document}

\maketitle

\begin{abstract}
In this paper, we propose surrogate agent-environment interface (SAEI) in reinforcement learning. We also state that learning based on probability surrogate agent-environment interface provides optimal policy of task agent-environment interface. We introduce surrogate probability action and develop the probability surrogate action deterministic policy gradient (PSADPG) algorithm based on SAEI. This algorithm enables continuous control of discrete action. The experiments show PSADPG achieves the performance of DQN in certain tasks with the stochastic optimal policy nature in the initial training stage.

\end{abstract}

\section{Introduction}

Reinforcement learning is an important topic in machine learning research. Its training relies on the interaction between agent and environment. With the development of artificial neural network, deep reinforcement learning is able to handle realistic real world problem.

Agent-environment interface describes the interaction between agent and environment in reinforcement learning. The boundary between agent and environment is preset according to task. "The agent-environment
boundary is determined once one has selected particular states, actions, and rewards, and thus has identified a specific decision-making task of interest" \citep{sutton_reinforcement_1998}. However, agent-environment interface is of little interest of algorithm development since it is related to the task definition itself more than how to solve the task.

In this paper we revisit the possibility of changing the interface in algorithm level and keep the interface intact in task level. Thus, we introduce a surrogate agent-environment interface. After introducing a surrogate probability action, we prove that the probability surrogate agent-environment interface gives the optimal policy solution to the task interface. In this framework, the learning agent interacts with surrogate agent-environment interface during the training process. It transforms the learned optimal policy of learning agent to the optimal policy of the task agent. To the best of our knowledge, it is the first time that a surrogate agent-environment interface is used to develop reinforcement learning algorithm. A search of the relevant literature yields little related articles. Some authors present surrogate action as an embedding vector in the continuous space for the original discrete actions \citep{dulac-arnold_deep_2015}. It does not change the interface during learning process. 

The contributions of this paper are as follow:
\begin{enumerate}
  \item It is the first time that the agent-environment interface is investigated for developing reinforcement learning algorithm.
  \item We prove that the surrogate probability agent-environment interface gives the optimal policy solution to the task interface.
  \item We develop the probability surrogate action deterministic policy gradient (PSADPG) algorithm based on SAEI which validate the surrogate agent-environment interface framework on algorithm development.
  \item PSADPG achieves the performance of DQN in certain tasks with the stochastic optimal policy nature in the initial training stage.
  \item PSADPG enables DQN \citep{mnih_human-level_2015} style off-policy learning algorithms (such as Double DQN \cite{van_hasselt_deep_2016}, Dueling DQN \citep{wang_dueling_2016}, Prioritized DQN \cite{schaul_prioritized_2015}) for stochastic discrete control. PSADPG augments the spectrum of deep reinforcement learning algorithm with extra dimensions.

\end{enumerate}

\section{Surrogate agent-environment interface}
\label{gen_inst}

The interaction between agent and environment is fundamental for reinforcement learning. In Figure 1, Agent performs action $a$ to environment. Environment dynamic then updates to next state and presents reward to agent. Markov decision process formally describes this interaction for reinforcement learning. A Markov decision process or a MDP consists of: set of states $S$, set of actions $A$, a probability function $P(s'|a,s)=P(s_{t+1}=s'|a_{t}=a,s_{t}=s)$ which gives the dynamic from state $s$ to state $s'$ under action $a$ at time $t$, a reward function $r_{t}=R(a,s)=R(a_{t}=a,s_{t}=s)$ which specifies the reward received at time $t$ after taking the action $a$ from state $s$. A policy for MDP is a function $a=\mu(s)$ or a probability distribution $\pi(a|s)$ determines an action $a$ in state $s$ at time $t$. The goal of reinforcement learning control is to search for policy that maximize the total reward $R=\sum_{t}\gamma^{t}r_{t}$ where $\gamma$ is a discount factor. 

\begin{figure}[h]
  \centering
  \fbox{
  \includegraphics[scale=0.5]{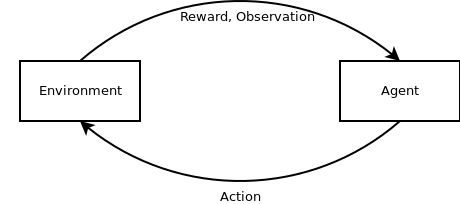} 
  }
  \caption{Task agent-environment interface}
\end{figure}

The deterministic policy directly gives a certain action. The stochastic policy, however, take additional sampling step after given a probability from distribution $\pi(a|s)$. Our idea is to extract this sampling step from agent and integrate it into environment. This presents a new interaction between agent and surrogate environment, see Figure 2. 

\begin{figure}[h]
  \centering
  \fbox{
  \includegraphics[scale=0.5]{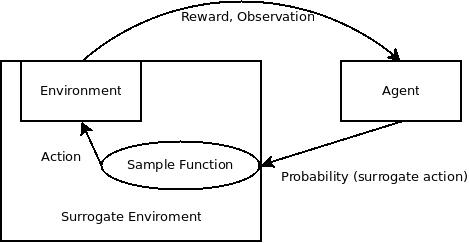} 
  }
  \caption{Surrogate agent-environment interface}
\end{figure}

In this setting, environment takes probability parameters as action from agent. The sampling process is part of environment. We do not assume agent has information of how the environment sample the action. To prove the feasibility of this framework, we have the following definitions. 

\newtheorem{Definition1}{Definition}
\newtheorem{Definition2}[Definition1]{Definition}

\begin{Definition1}
A task agent-environment interface $AEI_{t}$ is the agent-environment interface of the task of interest. A task Markov decision process $MDP_{t}$ is a MDP based on $AEI_{t}$. 
\end{Definition1}

\begin{Definition2}
The stochastic policy $\pi(s)$ of task agent-environment interface $AEI_{t}$ can be expressed as $\pi(s)=\phi\circ\widetilde{\mu}_{p}(s)$, where $\widetilde{\mu}_{p}(s)$ is a deterministic function mapping from state $s$ to action probability vector $p$ and $\phi$ is a sampling function mapping the probability vector $p$ to action $a$. For deterministic policy, $\mu(s)=\widetilde{\mu}_{p}(s)$. The function $\widetilde{\mu}_{p}(s)$ can be considered as a surrogate action policy from agent to environment and the sampling function $\phi$ is part of the surrogate environment. The function $\widetilde{\mu}_{p}(s)$ is called probability surrogate action policy. The resulting interface $\widetilde{AEI}_{p}$ is called probability surrogate agent-environment interface. The probability surrogate Markov decision process $\widetilde{MDP}_{p}$ is the $MDP$ based on $\widetilde{AEI}_{p}$.
\end{Definition2}

We prove that an optimal policy learned in probability surrogate agent-environment interface is equivalent to the optimal policy in task surrogate agent-environment interface.

\newtheorem{theorem1}{Theorem}
\begin{theorem1}
If the optimal probability surrogate policy in $\widetilde{MDP}_{p}$ is $\widetilde{\mu}_{p*}$. Then $\pi_{t*}=\phi\circ\widetilde{\mu}_{p*}$ is the optimal policy $\pi_{t*}$ in $MDP_{t}$ if the optimal policy is stochastic. If the optimal policy $\mu_{t*}$ in $MDP_{t}$ is deterministic, $\mu_{t*}=\widetilde{\mu}_{p*}$.
\end{theorem1}
\begin{proof}
In the case of stochastic policy, if $\pi_{t}'=\phi\circ\widetilde{\mu}_{p*}$ is not optimal in $MDP_{t}$, then there exists a state $s$ and policy $\pi$ such that $V_{\pi}(s)>V_{\pi_{t}'}(s)$. For $\pi(s)$ in $MDP_{t}$, there exists a $\widetilde{\mu}_{p}(s)$ in $\widetilde{MDP}_{p}$ such that $\pi(s)=\phi\circ\widetilde{\mu}_{p}(s)$. Since reward function is the same for both $MDP_{t}$ and $MDP_{t}$, $V_{\pi}(s)=V_{\widetilde{\mu}_{p}}(s)$ and $V_{\pi_{t}'}(s)=V_{\widetilde{\mu}_{p*}}(s)$. Thus, we have $V_{\widetilde{\mu}_{p}}(s)>V_{\widetilde{\mu}_{p*}}(s)$. This contradicts the optimality of $\widetilde{\mu}_{p*}$ in $\widetilde{MDP}_{p}$.
In the case of deterministic policy, $\mu_{t*}=\widetilde{\mu}_{p*}$ is trivial. 
\end{proof}

To validate the theorem, we introduce the probability surrogate action deterministic policy gradient (PSADPG) algorithm. This algorithm introduce a continuous approach on stochastic discrete action control to which off-policy policy gradient methods may apply. 

\section{Surrogate action deterministic policy gradient algorithm}
\label{headings}

The stochastic policy of discrete control not only gives a possible optimal solution but also enables a soft continuous learning process. Traditional policy gradient for discrete control utilizes likelihood ratio methods which incorporate probability distribution of action, e.g. REINFORCE \citep{williams_simple_1992}. The idea is to weight the probability of action by the reward based on this action. The state-of-the-art actor-critic algorithm A3C \citep{mnih_asynchronous_2016} is more along this line. 

Here we use a variant of deterministic policy gradient algorithm DPG \citep{silver_deterministic_2014} to directly capture the gradient of Q function respect to deterministic probability vector. DPG is specifically designed for continuous control. It handles the problem of instability of stochastic continuous policy gradient with the enhancement of efficiency. For high dimensional real world tasks, DDPG \citep{lillicrap_continuous_2015} is developed. With the probability surrogate action, we are able to transform the stochastic discrete control tasks into deterministic continuous control tasks. 

Algorithm 1 is a modified version of DDPG. For the purpose of comparison, we keep most of the symbols and statements intact from the original paper. Please refer to \citep{lillicrap_continuous_2015} for the detail of the algorithm. The difference from DDPG is that action $a$ in DDPG of learning process is replaced by probability vector $p$. Action $a$ in PSADPG sampled from $p$ is only used to interact with environment. Probability vector $p$ is the output of the softmax layer of actor network. To improve the efficiency of experience replay, we store a unit vector instead of the probability vector generated by the policy. The only non-zero element of this unit vector corresponds to the action $a_t$.

Since the algorithm is function approximation reinforcement learning approach, the optimality may not be guaranteed by the above theorem.

\begin{algorithm}
\caption{PSADPG Algorithm}\label{euclid}
\begin{algorithmic}
\STATE Randomly initialize critic network $Q(s,p|\theta^Q)$ and actor $\mu(s|\theta^\mu)$ with weights $\theta^Q$ and $\theta^\mu$.
\STATE Initialize target network $Q'$ and $\mu'$ with weights $\theta^{Q'} \leftarrow \theta^Q$ and $\theta^{\mu'} \leftarrow \theta^\mu$ 
\STATE Initialize replay buffer $R$
\FOR {episode = 1, M} 
	\STATE Receive initial observation state $s_1$
	\FOR {t=1, T} 
		\STATE Select probability $p_t=\mu(s_t|\theta^\mu)$ according to the current policy
		\STATE Sample action $a_t$ from probability $p_t$ with exploration
		\STATE Execute action $a_t$ and observe reward $r_t$ and observe new state $s_{t+1}$
		\STATE $\hat{p_t}=[\hat{p_t}_i]$ is the unit vector where $\hat{p_t}_i=1$ if $i=a_t$, $\hat{p_t}_i=0$ if $i\neq a_t$.
		\STATE Store transition $(s_t, \hat{p_t}, r_t, s_{t+1})$ in $R$
		\STATE Sample a random minibatch of $N$ transitions $(s_i, p_i, r_i, s_{i+1})$
		\STATE Set $y_i=r_i+\gamma Q'(s_{i+1}, \mu'(s_{i+1}|\theta^{\mu'})|\theta^{Q'})$
		\STATE Update critic by minimizing the loss: $L=\frac{1}{N} \sum_i(y_i-Q(s_i,p_i|\theta^Q))^2$
		\STATE Update the actor policy using the sampled policy gradient:
$$\bigtriangledown_{\theta^\mu}J\approx \frac{1}{N} \sum_i \bigtriangledown_p Q(s,p|\theta^Q)|_{s=s_i, p=\mu(s_i)} \bigtriangledown_{\theta^\mu}\mu(s|\theta^\mu)|_{s_i}$$
		\STATE Update the target networks:
$$\theta^{Q'} \leftarrow \tau\theta^Q + (1-\tau)\theta^{Q'}$$
$$\theta^{\mu'} \leftarrow \tau\theta^\mu + (1-\tau)\theta^{\mu'}$$
	\ENDFOR
\ENDFOR
\end{algorithmic}
\end{algorithm}

\section{Experiment}
\label{others}

To compare with the DQN algorithm, we test the PSADPG algorithm with DQN in classic discrete control. We also perform experiment in Atari 2600 game environment \citep{bellemare_arcade_2013}. We choose 'Acrobot-v1' and 'Amidar-v0' environment respectively in OpenAI gym \citep{brockman_openai_2016}. We select these two tasks for their stochastic optimal policy nature.

The classic control setup is as follow. For the critic network of classic control, state input is first embedded through a 64 units fully connected layer with hyperbolic tangent activation. The embedding vector is then linearly merged with probability input vector (with number of elements the same as number of actions) through 64 units fully connected layer. The critic network finally outputs the scalar Q value through another linear fully connected layer of 64 units. 

The actor network takes state input through a fully connected layers with 64 units with hyperbolic tangent activation. It then linearly outputs logits through a fully connected layer with number of units the same as number of actions. A softmax layer is used to output probability surrogate action. 

For the target network update, we use hard update instead of soft update which stated in the Algorithm 1. The update frequency is one update per 1000 iterations. 

Learning rate is set to be 0.0005. Gamma is 1.  Adam optimizer is used for stochastic gradient descend. The exploration is $\epsilon$ greedy with $\epsilon$ linearly reducing from 1 to 0.02 during first 100000 iterations and being constant as 0.02 thereafter.
 
For Atari game 'Amidar', the setup is almost the same. We first use the same convolutional network in DQN \citep{mnih_human-level_2015} to create input state. For all the fully connected layer, we use 512 units and rectified linear unit.

Figure 3 presents the learning curves of two algorithms on classic control task 'Acrobot'. Figure 4 shows the result for 'Amidar'. The episode reward is the mean value of the most recent 100 episodes. We can see two algorithms have similar learning curve performance in 'Acrobot'. In 'Amidar', for about 4000 episodes training, PSADPG performs more stable than DQN and almost achieves the result of DQN.

\begin{figure}[h]
  \centering
  \fbox{
  \includegraphics[scale=0.5]{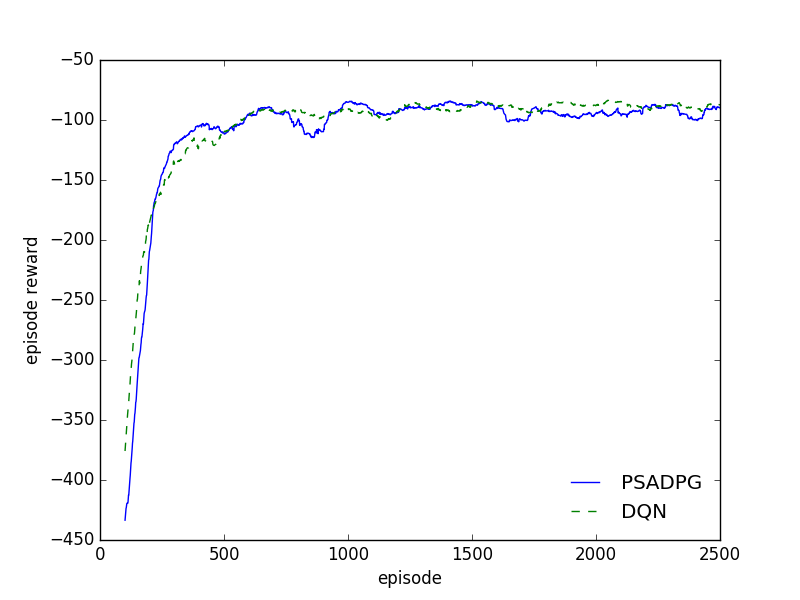} 
  }
  \caption{Learning curve of Acrobot-v1}
\end{figure}

\begin{figure}[h]
  \centering
  \fbox{
  \includegraphics[scale=0.5]{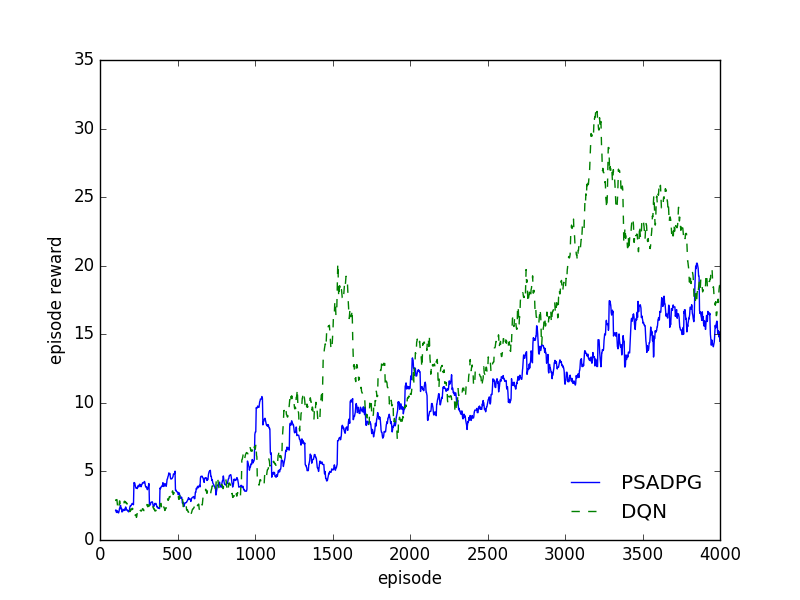} 
  }
  \caption{Learning curve of Amidar-v0}
\end{figure}

\section{Conclussion}

The surrogate agent-environment interface enables extra power to handle reinforcement learning tasks. In this paper, we prove that the policy optimality of probability surrogate agent-environment interface is equivalent to the task agent-environment interface. We also develop the algorithm to validate this theorem. The algorithm achieves the performance of DQN in certain tasks with the stochastic optimal policy nature in the initial training stage. We plan to explore more efficient algorithm based on SAEI in future work.

\bibliography{testpaper}

\begin{thebibliography}{10}

\bibitem{bellemare_arcade_2013}
Marc~G. Bellemare, Yavar Naddaf, Joel Veness, and Michael Bowling.
\newblock The {Arcade} {Learning} {Environment}: {An} evaluation platform for
  general agents.
\newblock {\em J. Artif. Intell. Res.(JAIR)}, 47:253--279, 2013.

\bibitem{brockman_openai_2016}
Greg Brockman, Vicki Cheung, Ludwig Pettersson, Jonas Schneider, John Schulman,
  Jie Tang, and Wojciech Zaremba.
\newblock {OpenAI} {Gym}.
\newblock {\em arXiv preprint arXiv:1606.01540}, 2016.

\bibitem{dulac-arnold_deep_2015}
Gabriel Dulac-Arnold, Richard Evans, Hado van Hasselt, Peter Sunehag, Timothy
  Lillicrap, Jonathan Hunt, Timothy Mann, Theophane Weber, Thomas Degris, and
  Ben Coppin.
\newblock Deep {Reinforcement} {Learning} in {Large} {Discrete} {Action}
  {Spaces}.
\newblock {\em arXiv:1512.07679 [cs, stat]}, December 2015.
\newblock arXiv: 1512.07679.

\bibitem{lillicrap_continuous_2015}
Timothy~P. Lillicrap, Jonathan~J. Hunt, Alexander Pritzel, Nicolas Heess, Tom
  Erez, Yuval Tassa, David Silver, and Daan Wierstra.
\newblock Continuous control with deep reinforcement learning.
\newblock {\em arXiv preprint arXiv:1509.02971}, 2015.

\bibitem{mnih_asynchronous_2016}
Volodymyr Mnih, Adria~Puigdomenech Badia, Mehdi Mirza, Alex Graves, Timothy
  Lillicrap, Tim Harley, David Silver, and Koray Kavukcuoglu.
\newblock Asynchronous methods for deep reinforcement learning.
\newblock In {\em International {Conference} on {Machine} {Learning}}, pages
  1928--1937, 2016.

\bibitem{mnih_human-level_2015}
Volodymyr Mnih, Koray Kavukcuoglu, David Silver, Andrei~A. Rusu, Joel Veness,
  Marc~G. Bellemare, Alex Graves, Martin Riedmiller, Andreas~K. Fidjeland,
  Georg Ostrovski, and {others}.
\newblock Human-level control through deep reinforcement learning.
\newblock {\em Nature}, 518(7540):529--533, 2015.

\bibitem{schaul_prioritized_2015}
Tom Schaul, John Quan, Ioannis Antonoglou, and David Silver.
\newblock Prioritized experience replay.
\newblock {\em arXiv preprint arXiv:1511.05952}, 2015.

\bibitem{silver_deterministic_2014}
David Silver, Guy Lever, Nicolas Heess, Thomas Degris, Daan Wierstra, and
  Martin Riedmiller.
\newblock Deterministic policy gradient algorithms.
\newblock In {\em Proceedings of the 31st {International} {Conference} on
  {Machine} {Learning} ({ICML}-14)}, pages 387--395, 2014.

\bibitem{sutton_reinforcement_1998}
Richard~S. Sutton and Andrew~G. Barto.
\newblock {\em Reinforcement learning: {An} introduction}, volume~1.
\newblock MIT press Cambridge, 1998.

\bibitem{van_hasselt_deep_2016}
Hado Van~Hasselt, Arthur Guez, and David Silver.
\newblock Deep {Reinforcement} {Learning} with {Double} {Q}-{Learning}.
\newblock In {\em {AAAI}}, pages 2094--2100, 2016.

\bibitem{wang_dueling_2016}
Ziyu Wang, Tom Schaul, Matteo Hessel, Hado Hasselt, Marc Lanctot, and Nando
  Freitas.
\newblock Dueling {Network} {Architectures} for {Deep} {Reinforcement}
  {Learning}.
\newblock In {\em International {Conference} on {Machine} {Learning}}, pages
  1995--2003, 2016.

\bibitem{williams_simple_1992}
Ronald~J. Williams.
\newblock Simple statistical gradient-following algorithms for connectionist
  reinforcement learning.
\newblock {\em Machine learning}, 8(3-4):229--256, 1992.

\end{thebibliography}
\bibliographystyle{plain}

\end{document}